\documentclass[10pt,journal,twoside,twocolumn,romanappendices]{IEEEtran}

\usepackage{psfrag,epsfig,graphics}
\usepackage{amsmath,amssymb,amsthm,multirow}
\usepackage{cite,subfigure,balance}
\usepackage{url}
\usepackage[justification=centering]{caption}

\usepackage{color}
\def\cred{\textcolor{black}}  

\definecolor{violet}{rgb}{0.5,0,0.5}

\newcommand{\cb}[1]{{\boldsymbol{#1}}}
\newcommand{\cp}[1]{\ifmmode {\mathcal{#1}}\else ${\mathcal{#1}}$\fi}
\newcommand{\balpha}{\boldsymbol{\alpha}}

\newcommand{\bM}{\boldsymbol{M}}

\newcommand{\bX}{\boldsymbol{X}}

\newcommand{\bm}{\boldsymbol{m}}
\newcommand{\bn}{\boldsymbol{n}}

\newcommand{\br}{\boldsymbol{r}}

\newcommand{\bx}{\boldsymbol{x}}

\usepackage{tikz}

\begin{document}

\title{A laboratory-created dataset with ground-truth for hyperspectral unmixing evaluation}

\author{Min Zhao,  \IEEEmembership{Student Member, IEEE},  Jie Chen, \IEEEmembership{Senior Member, IEEE},  Zhe He, \IEEEmembership{Member, IEEE}
\thanks{The work of J. Chen was supported in part by Natural Science Foundation of Shenzhen under grant JCYJ2017030155315873. A short and preliminary version of this work appears in the IGARSS conference publication~\cite{Zhao2018}.}
\thanks{M. Zhao and J. Chen are with School of Marine Science and Technology, Northwestern Polytechnical University, China. Zhe He is with Department of Geomatics Engineering, University of Calgary, Canada. (corresponding author: J. Chen, dr.jie.chen@ieee.org).}
\thanks{Dataset is available on corresponding author's website: http://www.jie-chen.com/.}
}

\maketitle

\begin{abstract}
Spectral unmixing is an important and challenging problem in hyperspectral data processing. This topic has been extensively studied and a variety of unmixing algorithms have been proposed in the literature. However, the lack of publicly available dataset with ground-truth makes it difficult to evaluate and compare the performance of unmixing algorithms in a quantitative and objective manner. Most of the existing works rely on the use of numerical synthetic data and an intuitive inspection of the results of real data. To alleviate this dilemma, in this study, we design several experimental scenes in our laboratory, \cred{including printed checkerboards, mixed quartz sands, and reflection with a vertical board.} A dataset is then created by imaging these scenes with the hyperspectral camera in our laboratory, providing \cred{$36$} mixtures  with more than \cred{$130, 000$} pixels with $256$ wavelength bands ranging from $400nm$ to $1000nm$. The experimental settings are strictly controlled so that pure material spectral signatures and material compositions are known. To the best of our  knowledge, this dataset is the first publicly available dataset created in a systematic manner with ground-truth for spectral unmixing. Some typical linear and nonlinear unmixing algorithms are also tested with this dataset and lead to meaningful results.
\end{abstract}


\section{Introduction}

Hyperspectral imaging is a continuously growing field of study and has received considerable attention in the last decade.  Hyperspectral data provide a wide spectral range, coupled with a high spectral resolution. These characteristics are suitable for the detection and classification of surfaces and chemical elements in the observed images. Rich information in the spectral dimension provides solutions to many problems that cannot be solved by traditional RGB imaging or multispectral imaging. However, because of factors such as the low spatial resolution of the spectral imaging devices, and the diversity of materials in a scene, an observed pixel may contain several materials. Spectral unmixing plays an important role in hyperspectral data processing, aimed at separating a pixel spectrum into a set of spectral signatures, termed endmembers, and a set of fractional abundances associated with each endmember~ \cite{bioucas2012hyperspectral}.

A significant amount of effort has been made in the past decade to solve the spectral unmixing problem. Endmember identification and abundance estimation can be conducted either in a sequential or a simultaneous manner. Sequential techniques first determine the endmembers, endmember extraction algorithms such as purity index algorithm (PPI)~\cite{boardman1993automating}, vertex component analysis (VCA)~\cite{nascimento2005does} and N-FINDR algorithm~\cite{winter1999n}. In the sequel, the problem boils down to determination of the fractional abundances~\cite{heize2001fully}. Simultaneous techniques determine the endmembers and the estimated abundance fractions at the same time, mainly by using matrix decomposition methods~\cite{miao2007endmember},  sparse learning methods~\cite{Iordache2011}, or optimization methods~\cite{Li2015} to this purpose. Moreover, several other facts have been further taken into account to enhance the modeling capacity and improve the performance of the unmixing algorithms. As hyperspectral images inherently contain two-dimensional information, it is natural to impose the spatial continuity on the estimated abundances~\cite{Iordache2012,Chen2014}.  Considering the limitation of the linear unmixing due to the multilayer reflectance and the intimate interaction of materials, several works have investigated the nonlinear models and nonlinear unmixing algorithms~\cite{halimi2011nonlinear,chen2013nonlinear,qu2014abundance,tang2017integrating}. Taking into account the endmember variability in the observed scenario,  researchers have considered a bundle of spectra to be signatures of a material during the unmixing process~\cite{bateson2000endmember,somers2012automated,xu2015image}.

In contrast to the increasing development of various types of unmixing algorithms, the way to perform an effective and objective evaluation of these algorithms remains restrictive and unconvincing, although both synthetic data and real data are used in the literature. Compared with the target detection and classification tasks where several labeled datasets are available, the lack of ground-truth information for the unmixing makes it difficult to evaluate and compare the algorithms in an objective manner.  On one hand, synthetic data are generated by the assumed mathematical models. Validating an algorithm with the data generated by the same model restricts the evaluation of the general performance of the algorithm. On the other hand, as there is no ground-truth dataset for the spectral unmixing task, it is difficult to interpret the unmixing results of the algorithms even though they have been applied to real datasets. This restricts a comparison of the results on the basis of the observation and the intuitive interpretation of the estimated abundance maps.  See Sec.~\ref{sec:evap} for a more detailed review of and discussion on this issue.


Many hyperspectral datasets are captured by airborne devices and satellites. Labeling these data with material distribution fractions is extremely difficult. In this work, we design several experimental scenes in our laboratory with controlled settings, where the pure material spectra and the material composition methods are known. We then capture these scenes with a hyperspectral device and perform the necessary preprocessing to generate the dataset. To the best of our  knowledge, this dataset is the first publicly available dataset with ground-truth for spectral unmixing created in a systematic manner. A summary of our contributions and the main properties of the data as follows:
\begin{itemize}
     \item We build a laboratory-created dataset with ground-truth for evaluating hyperspectral unmixing algorithms. This dataset is available on the corresponding author's website http://www.jie-chen.com/.
     \item We design several scenes to model different mixture mechanisms, \cred{including printed checkerboards, mixed quartz sands, and reflection with a vertical board.} Both the images of pure materials and the composition of each material are provided so that researchers have access to the endmembers and the corresponding abundances. The endmember variability and nonlinear mixture effects can also be observed with the created data. See Sec. III for further details on the settings.
     \item We apply several typical unmixing techniques to these data. We find that linear unmixing algorithms result in abundance RMSEs of the order of $10^{-2}$ for checkerboard-type data, and the nonlinear unmixing algorithms such as K-Hype and Hapke are beneficial for intimately mixed materials.
\end{itemize}
Note that this laboratory-created dataset does not have the exact same properties as real airborne and satellite data. However, it is expected to help in reducing  the gap between the theoretical research and practice, and to help researchers to evaluate their unmixing algorithms in a quantifiable and comparable manner.

The rest of this paper is organized as follows. Section II reviews several popular spectral mixing models, and presents the  status and limitations of the current methods of evaluating the unmixing performance. Section III presents the properties, capturing settings, and designs scenes of our created dataset in detail. In order to validate our dataset, we apply several typical endmember extraction algorithms and unmixing techniques to the created scenes; this is described in Section IV.  Section V concludes the paper and presents the discussion.

\bigskip

\noindent\textbf{Notation.} Normal font $x$ and $X$ denote scalars. Boldface small letters $\bx$ denote vectors. All vectors are column vectors. Boldface capital letters $\bX$ denote matrices.  Specifically, let $\br$ be an observed column pixel, supposed to be a mixture of $R$ endmember spectra, with $L$ the number of spectral bands. Assume that $\bM = [\bm_1, \bm_2, \dots, \bm_R]$ is the  $(L\times R)$   target endmember matrix, where each column $\bm_i$ is an endmember spectral signature. For the sake of convenience, we shall denote by $\bm_{\lambda_\ell}^\top$ the $\ell$-th $(1\times R)$ row of $\bM$,  that is, the vector of the $R$ endmember signatures at the $\ell$-th wavelength band. Let $\balpha = [\alpha_1, \alpha_2, \dots, \alpha_R]^\top$ be the abundance column vector associated with the pixel .

\section{Existing mixing models and unmixing performance evaluation}
In this section, we will briefly review several widely used mixing models and summarize the usual methods of evaluating the unmixing performance in the literature. We shall see that these models and restrictions in the existing evaluation methodology considerably motivate the experimental settings in this work.
\subsection{Widely used mixture models}
\subsubsection{Linear mixture model}
The linear mixture model (LMM) is widely used to identify and quantify pure components in remotely sensed images due to its simple physical interpretation and trackable estimation process. If the multiple scattering among distinct endmembers is negligible  and distinct materials are considered to be partitioned in a checkerboard manner,  the spectrum of a low resolution pixel is  approximated by a linear mixture of endmember spectra weighted by the corresponding fractional abundances\cite{bioucas2012hyperspectral}\cite{keshava2002spectral}\cite{eches2010bayesian}, namely, the observed $\br$, is given by
\begin{equation}
           \br = \bM \balpha + \bn
\end{equation}
with $\bn \in \mathbb{R}^L$ being a modeling noise. Specifically, since components of $\balpha$ represent fractions of each material, it is often considered that they satisfy the non-negativity (ANC) and sum-to-one constraints (ASC):
\begin{align}
             &\alpha_i \geq 0,  \qquad  \forall i\in \{1, \dots, R\} \\
             \sum_{i=1}^R &\alpha_i = 1.
\end{align}
Although the linear mixture model has obvious practical advantages, there are many situations in which it may not be appropriate and can be advantageously replaced by a nonlinear one. Some typical ones are presented here-below.
\subsubsection{Bilinear mixing model}
Considering that multiple scattering effects can be observed on complex vegetated surfaces where it is assumed that incident solar radiation is scattered through multiple bounces involving several endmembers. Bilinear model accounts for presence of second-order photon interactions by introducing additional interaction terms in the linear model.  The model is given by:
\begin{equation}
           \br = \bM \balpha + \sum_{i=1}^{R-1}\sum_{j=i+1}^R \beta_i \beta_j (\bm_i \otimes \bm_j) + \bn
\end{equation}
where $\otimes$ denotes the element-wise product of two vectors. Relating $\beta_i$ and $\balpha$ with different constraints leads a variety of bilinear models~\cite{halimi2011nonlinear}.
\subsubsection{Linear mixing/nonlinear fluctuation model}
In our previous work\cite{chen2013nonlinear}, we do not restrict to use the extra polynomial terms to describe the nonlinearity as in the bilinear model. We  assume that the mixing mechanism can be described by a linear mixture of endmember spectra, with additive nonlinear fluctuations that allows to take complex interactions of endmembers into account.
\begin{equation}
        \label{eq:model.linf}
           [\br]_{\ell} =  \bm_{\lambda_{\ell}}^\top \,  \balpha + \psi_{\rm nlin} (\bm_{\lambda_{\ell}}) + [\bn]_\ell
\end{equation}
where $\psi_{\rm nlin}$ can be any real-valued functions on a compact $\cp{M}$, of a reproducing kernel Hilbert space with $\kappa_{\rm nlin}$ being its reproducing kernel. Selecting  kernels allows to capture the  different types of nonlinearity of the mixture model.  Performing the unmixing with~\eqref{eq:model.linfa} suggests to simultaneously estimate the abundance $\balpha$ and the nonlinear fluctuation $\psi_{\rm nlin}$ with the so-called kernel trick. Introducing the abundance $\balpha$ in the nonlinear term with a post-nonlinear form, we further have
\begin{equation}
        \label{eq:model.linfa}
           [\br]_{\ell} =  \bm_{\lambda_{\ell}}^\top \,  \balpha + \psi_{\rm nlin} ( \bm_{\lambda_{\ell}}^\top \,  \balpha ) + [\bn]_\ell.
\end{equation}
This model was proposed in\cite{chen2013estimating}. In\cite{altmann2014residual}\cite{altmann2015bayesian}, the term $\psi_{\rm nlin}$ is also called the residual term and the associated algorithm is called the residual component analysis.
\subsubsection{Intimate model}
Intimate mixtures are mixtures of grains or particles that are in close contact with each other. Light  typically interacts multiple times with the particles making up the mixture before reaching the observer.  The optical characteristics of such mixtures depend on many parameters, such as the total number and the fractions of each component, the grain size distribution of each component, the shape and orientation of the grains, the average optical distance between reflections, the absorption and scattering characteristics, etc. Model proposed by Hapke~\cite{hapke1981bidirectional} is a popular model that describes the optical characteristics of intimately mixed materials. It can be viewed as a post-nonlinear model applied to a linear mixture in the albedo domain. See~\cite{hapke1981bidirectional} for more details. Though the intimate model has a clear physical interpretation, few works perform unmixing with this model due to its complexity and requirement of several parameters of the scene.

\subsection{Evaluation of the unmixing performance}
\label{sec:evap}
In contrast to the increasing development of  various types of unmixing algorithms, the way to perform an effective and objective evaluation of algorithms remains restrictive and unconvincing. Most works in the literature evaluate the unmixing performance with both synthetic data and real data, however, with significant limitations.
\subsubsection{Evaluation with synthetic data} Synthetic data are generated using an assumed mathematical model.  The use of  synthetic data is considered as a baseline to evaluate and compare the unmixing algorithms, as all information, including endmembers, abundances, and the mixing model, are known. Therefore we can evaluate  algorithms with objective measures. However, it is clear that the use of synthetic data is very restrictive because of the following reasons:
\begin{itemize}
        \item Each model has its own limitation in modeling a real mixture in the observed scene. It is not trivial to predict to the performance of an algorithm (particularly nonlinear ones) in real applications, even though it works well with the synthetic data generated using a mathematical model.
        \item It is common in the exiting literature to validate an algorithm with data generated by the same model. It is not convincing to evaluate the performance of an algorithm in such a manner. For instance, an algorithm carefully devised from the bilinear model may outperform other nonlinear unmixing algorithms when we use data generated by the same model, however, it does not make much sense if the performance of this algorithm is severely biased when the practical mixture is not ideally bilinear.
\end{itemize}
\subsubsection{Evaluation with real data} Besides using synthetic data, most of the previous studies also \cred{use} real images such as the Cuprite dataset~\cite{chen2013nonlinear,eches2010bayesian,qu2014abundance,Yang2016,Sigurdsson2016} to validate the proposed algorithm.  However, people are currently faced with a dilemma of using these real datasets captured by airborne methods.  On one hand, while considering  the limitation of synthetic data, evaluating algorithms with real data is indeed important. On the other hand, it is very difficult to compare the obtained results because the lack of the associated ground-truth information for real data precludes a quantitative and objective evaluation of algorithms. Researchers make the following efforts to evaluate the unmixing performance while using real data:
\begin{itemize}
      \item Interpreting results with intuitive observation: When testing unmixing algorithms with real data, most of the existing works~\cite{chen2013estimating,eches2010bayesian,halimi2011nonlinear,Giampouras2016,Sigurdsson2016,Yang2016,Halimi2017,Jiang2018} illustrate the results in a straightforward manner by showing the estimated abundance maps and making visually intuitive comparisons. Clearly, this qualitative method is not very helpful to understand the performance of these algorithms.
       \item Comparing  reconstruction errors: Besides an intuitive examination of  abundance maps, the spectral reconstruction error (RE) and the spectral angle (SA) between the observed pixel and the reconstructed one are frequently used as quantitative measures in the literature~\cite{halimi2011nonlinear,altmann2015bayesian,Chen2014,Halimi2017}. However, it is widely accepted that the quality of reconstruction is not necessarily in proportion to the quality of unmixing, particularly for real images where a complex nonlinear mixture is involved~\cite{chen2013nonlinear,altmann2014residual,qu2014abundance}.
      \item Comparing  results with classification results: In contrast to the unmixing task, there exist several datasets with ground-truth labels for the hyperspectral classification. Some works benefit from this information to evaluate the unmixing result in an indirect manner. In~\cite{Dopido2011,Chen2014}, the authors  consider the  abundance obtained from unmixing algorithms as the feature of a classifier, and then compare the classification results with the ground-truth. In~\cite{Iordache2012,Iordache2013,Akhtar2017,Zhang2018,Jiang2018}, the authors compare the abundance maps with the classification results obtained using the Tricorder software.
      \item Comparing  results with those obtained from existing algorithms: A handful of works consider the results from a given unmixing algorithm as the standard reference, and then compare the performance of other algorithms with it. For instance, in~\cite{shi2016linear}, the authors use this methodology to generate ``labeled" unmixing data. In~\cite{Qian2011,lu2013manifold}, the authors evaluate the results of NMF by comparing them with those of FCLS.
\end{itemize}
It is clear that none of the above methods provide a convincing quantitative performance evaluation of unmixing algorithms. The result illustration with real data in the existing works is therefore not highly informative and persuasive. This motivates us to create a proper dataset with ground-truth in our laboratory for a fair evaluation of unmixing algorithms.

\begin{figure*}[!th]
  \centering
  \centerline{\includegraphics[width=16cm]{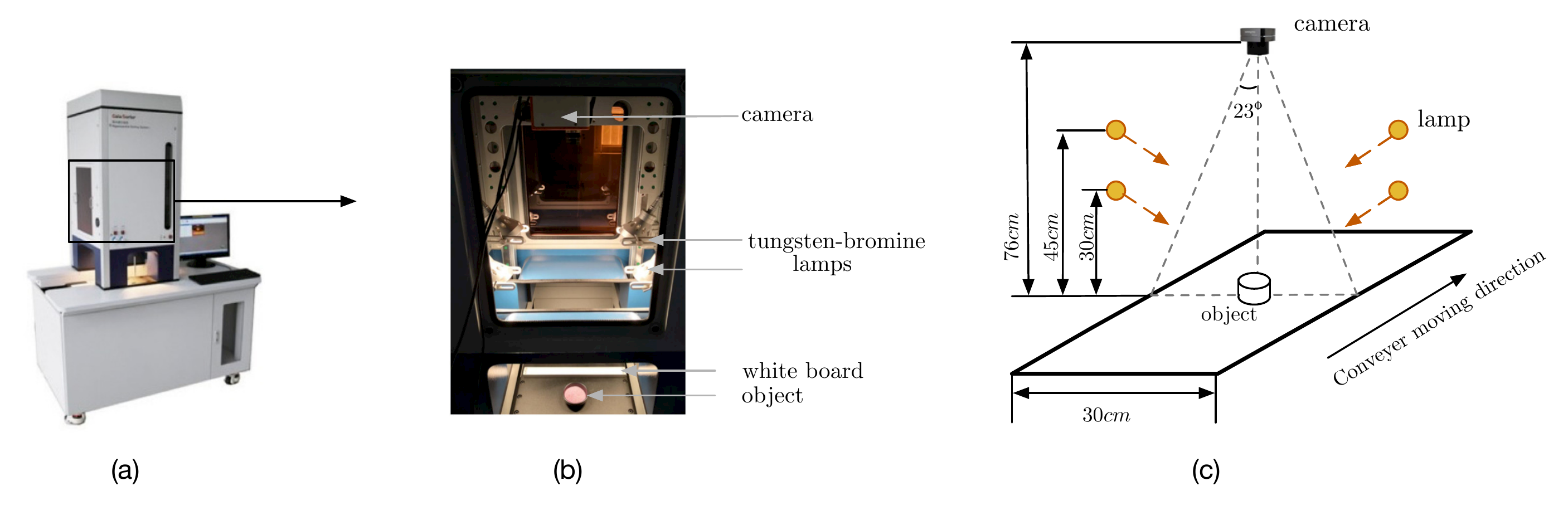}}
  \caption{Data capture device and schema: (a) Imaging device (Gaia Sorter). (b) Details in the imaging obscura. (c) A diagram of data acquisition setting.}
\label{fig:camera}
\end{figure*}

\section{Description of the dataset}
\subsection{General description and properties of the dataset}
Our dataset possesses the following properties:
\begin{itemize}
\item A variety of mixing scenes: We design several different scenes to mimic various mixture models with experiment parameters being strictly controlled, \cred{including checkerboard-type data, mixed quartz sands, and reflection with a vertical board.} A single scene consists of up to 4 materials and 14 different mixtures.
\item Available ground-truth: Both material endmembers and their associated abundances are known. We image pure materials for collecting the endmembers, and use the prior known compositions {or high-resolution images} to provide the abundance fractions.  With this information, both of the endmember extraction and the abundance estimation algorithms can be evaluated.
\item Spatial property: A raw image has a spatial resolution of $456\times348$ pixels. In order to provide data with uniform illumination and remove  unrelated objects (e.g., conveyer background, calibration board, the container), we keep the region of $60\times60$ pixels from the center of each scene. Both raw images with considerably more pixels and the clipped images are provided in our dataset, where the clipped images already provide at least 3600 pixels with ground-truth.
\item Spectral property: The data have 256 channels in the spectral dimension, ranging from $400 nm$ to $1000 nm$, with a spectral resolution of up to $0.58nm$. The noise of each band is not uniform. First and last several bands are much noisy due to the low sensitivity of the spectrometer at these two ends. These bands can simply be discarded, or kept for studying the band selection, robust unmixing, etc.

\end{itemize}
\subsection{Methods}
\subsubsection{Hyperspectral device description and settings}
 Our data were collected by the GaiaField and GaiaSorter systems in our laboratory. Our GaiaField (Sichuan Dualix Spectral ImageTechnology Co. Ltd., GaiaField-V10)  is a push-boom imaging spectrometer with a HSIA-OL50 lens, covering the visible and NIR wavelengths ranging from $400 nm$ to $1000 nm$, with a spectral resolution of up to $0.58nm$.
 This camera adopts CCD instead of CMOS detector, leading to a lower noise level. Provided that the camera is not running in an environment with a very high temperature, the heating effect, readout noise and dark current noise will be stable. The dark current has a negligible effect for the visible and NIR range.
GaiaSorter sets an environment that isolates external lights, and is endowed with a conveyer to move samples for the push-boom imaging.
 Conveyor belt is made of aluminum with the oxidizing blackened surface. It can be considered as a black-body in the most range of $400-1000 nm$.
 Four tungsten-bromine lamps were used to form  a hemispherical-directional illumination that simulated uniform real-world solar illumination.
 A halogen lamp is characterized by its stability in illumination. After a warm-up time of $15 min$, the spatial illumination uniformity of our lamps is greater than $90\%$ and the light sources are proved to be stable over time.
 The distance between the camera and the samples was adjustable. The exposure time, moving speed of the conveyer, and focusing were automatically computed and controlled by the associated software that was run on a computer.

In our experiments, the camera was right above the target. The distance between the lens and the samples was adjusted to $76 cm$, and the field angle of the camera was {$23^{\circ}$}. Therefore, the width of view was $30 cm$. The vertical distance between the lamps and the samples was approximately $30 cm$. At these settings, a collected raw hyperspectral cube consisted of  348 and 455 pixels in width and height, and 256 spectral bands. We then found that the resolution of a collected image is $0.86 mm$/pixel. In our experiments, we controlled the cross-sectional areas of each pure material to be less than $0.86 mm\times0.86 mm$, to ensure that each pixel was a mixture of multiple materials.

\subsubsection{Data normalization}
In order to convert the collected light intensity levels to reflectance values, we applied the black-white normalization to the raw data. This normalization also aimed to remove the effect of the dark current of the camera sensor and avoided the uneven light intensity of each band. In the offline phase, the black image is acquired by turning off the light source and covering the camera lens with its cap. The white image was acquired by imaging a standard white board (foamed PEFE resin, certified by China Metrology Institute) under the same condition as that of the raw image for calibration. Then, a normalized pixel $\br_{\rm norm}$ was obtained by performing the following step on the raw observation $\br_{\rm raw}$~\cite{serranti2015hyperspectral}:
\begin{equation}
      \label{eq:bwnorm}
         \br_{\rm norm} = \frac{\br_{\rm raw} - \br_{\rm black}}{\br_{\rm white} - \br_{\rm black}}
\end{equation}
where $\br_{\rm black}$ and $\br_{\rm white}$ denote the spectra of the black and white frames respectively, and the difference and the division operations are applied in an element-wise manner.
\subsection{Experimental scenes and data description}
\subsubsection{Scene \uppercase\expandafter{\romannumeral1}}
\begin{figure*}[!th]
  \centering
  \centerline{\includegraphics[width=15cm]{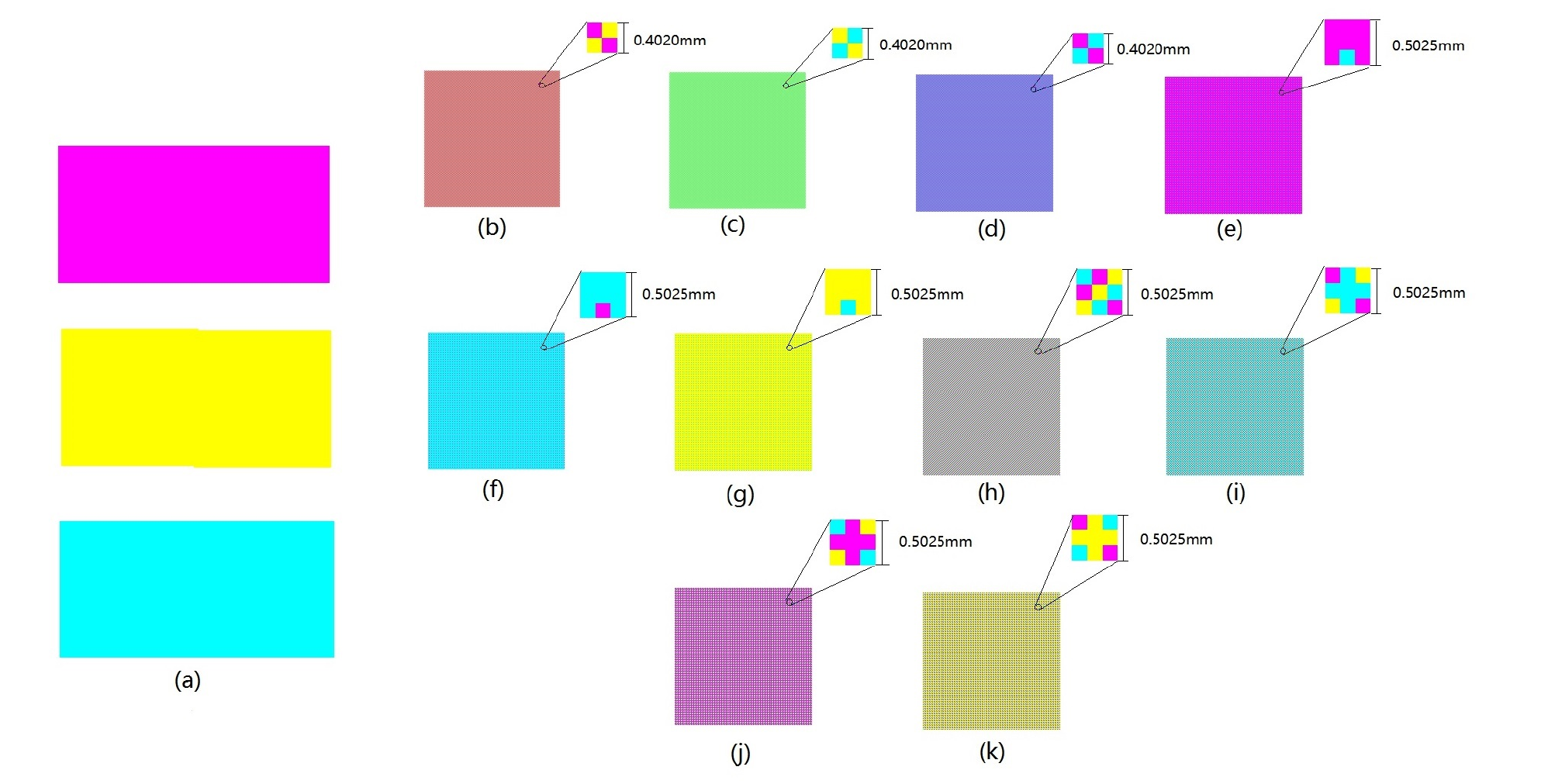}}
  \caption{Illustration of Scene \uppercase\expandafter{\romannumeral1}. (a): Three pure color blocks (magenta, yellow  and cyan). (b) to (k): Checkerboards consisting of ${0.2010 mm\times0.2010 mm}$ or ${0.1675 mm\times0.1675 mm}$ squares with different layouts to form ten mixtures. Zoomed-in part of each subfigure illustrate the sizes and arrangements of color blocks.}
\label{fig:scene1}
\end{figure*}

\begin{figure*}[!th]
  \centering
  \centerline{\includegraphics[width=19cm]{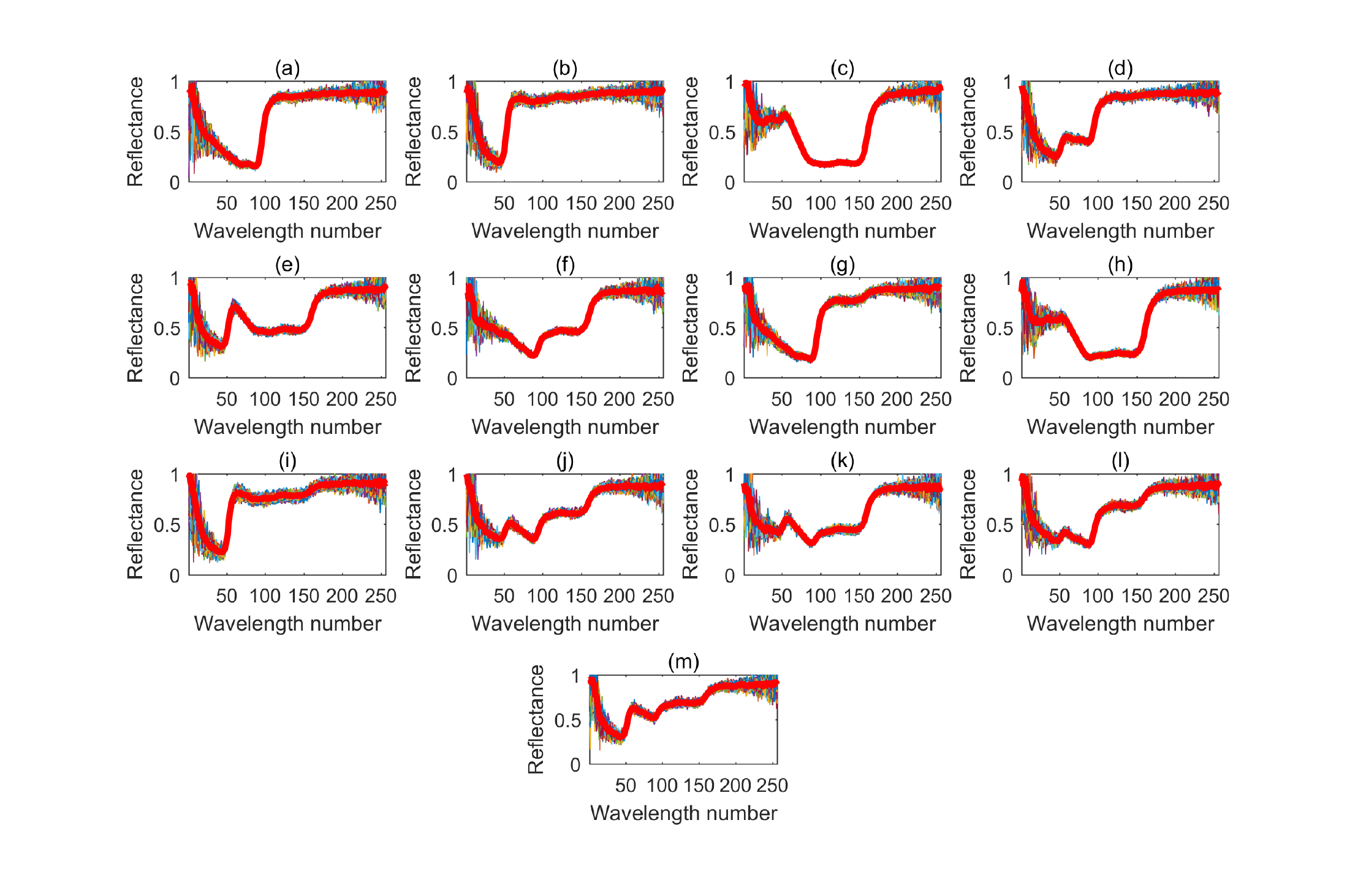}}
  \vspace{-1cm}
  \caption{Spectral curves of Scene \uppercase\expandafter{\romannumeral1}. Spectra of 50 pixels are plotted for each case, and the red curve represents the mean spectrum. (a) to (c): Spectra of pure color blocks, magenta, yellow, cyan in turn (endmembers). (d) to (m): Spectra of ten mixtures, Mixture 1- Mixture 10 in turn.}
\label{fig:curves1}
\end{figure*}

In the first scene, we aimed to design a scene that matched the linear mixing model. As in the linear model, the abundance can be seen as the area percentage of each material in a pixel; we used a checkerboard image to set this scene. Three typical colors of color printing, namely, magenta, yellow, and cyan, were used as the endmembers, as shown in Fig.~\ref{fig:scene1}(a).
We printed these colors in a checkerboard manner consisting of a large number of  ${0.2010 mm\times0.2010 mm}$ or ${0.1675 mm\times0.1675 mm}$ squares.
The reference materials and the mixed scenes are printed on the same type of paper.
This setting mimicked the low resolution data capture process and was consistent with the assumption of the linear mixture model. We prepared ten mixture settings in this way with the fraction abundances reported in Tab.~\ref{tab:standard1}, and the mixture patterns  depicted pictorially in Figs.~\ref{fig:scene1}(b-k). The spectral curves of the endmembers and the mixtures are illustrated in Fig.~\ref{fig:curves1}.
\begin{table} [t]
   \footnotesize \centering
   \caption{\small Abundance ground-truth of Scene \uppercase\expandafter{\romannumeral1}.}
    \vspace{-1mm}
  \begin{tabular}{c|ccc}
  \hline\hline
     & Magenta & Yellow & Cyan \\
  \hline
  Mixture 1 & $50.00\%$ & $50.00\%$ & 0\\
  Mixture 2 & 0& $50.00\%$ & $50.00\%$\\
  Mixture 3 & $50.00\%$ & 0 & $50.00\%$\\
  Mixture 4 & $88.89\%$ & 0 & $11.11\%$\\
  Mixture 5 & $11.11\%$ & 0 & $88.89\%$\\
  Mixture 6 & 0 & $88.89\%$ & $11.11\%$\\
  Mixture 7 & $33.33\%$ & $33.33\%$ & $33.33\%$\\
  Mixture 8 & $22.22\%$ & $22.22\%$ & $55.56\%$\\
  Mixture 9 & $55.56\%$ & $22.22\%$ & $22.22\%$\\
  Mixture 10 & $22.22\%$ & $55.56\%$ & $22.22\%$\\
  \hline\hline
  \end{tabular}
\label{tab:standard1}
  \end{table}

\subsubsection{Scene \uppercase\expandafter{\romannumeral2}}
\begin{figure*}
  \centering
  \centerline{\includegraphics[width=16cm]{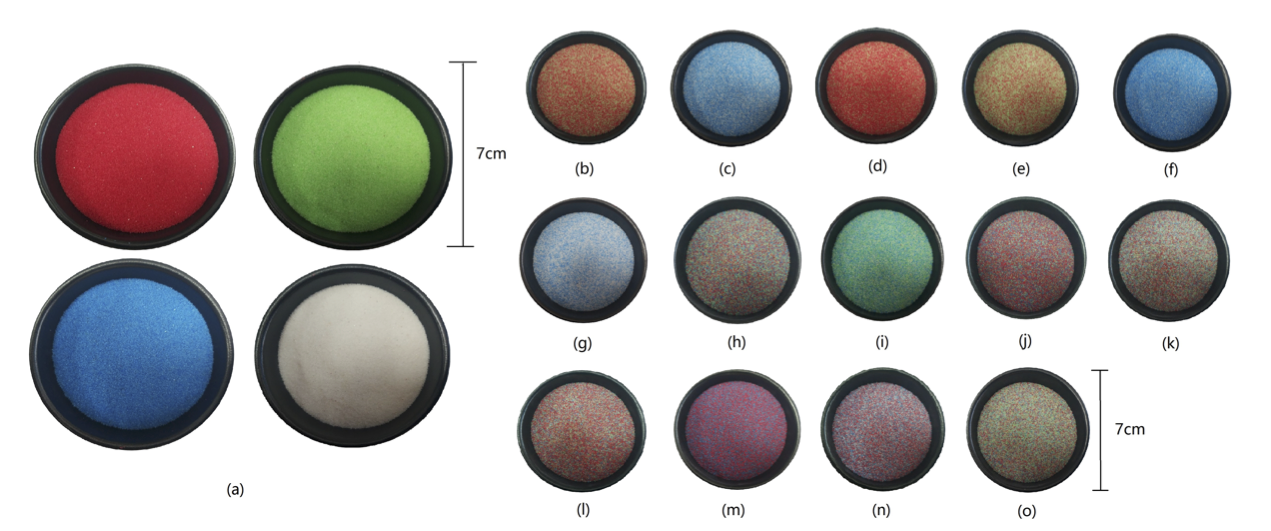}}
  \caption{Diagram of Scene \uppercase\expandafter{\romannumeral2} with uniform mixtures, (a) Pure colored quartz sands. (b)-(o) Fourteen intimate mixtures.}
\label{fig:scene2}
\end{figure*}

\begin{figure*}
  \centering
  \centerline{\includegraphics[width=20cm]{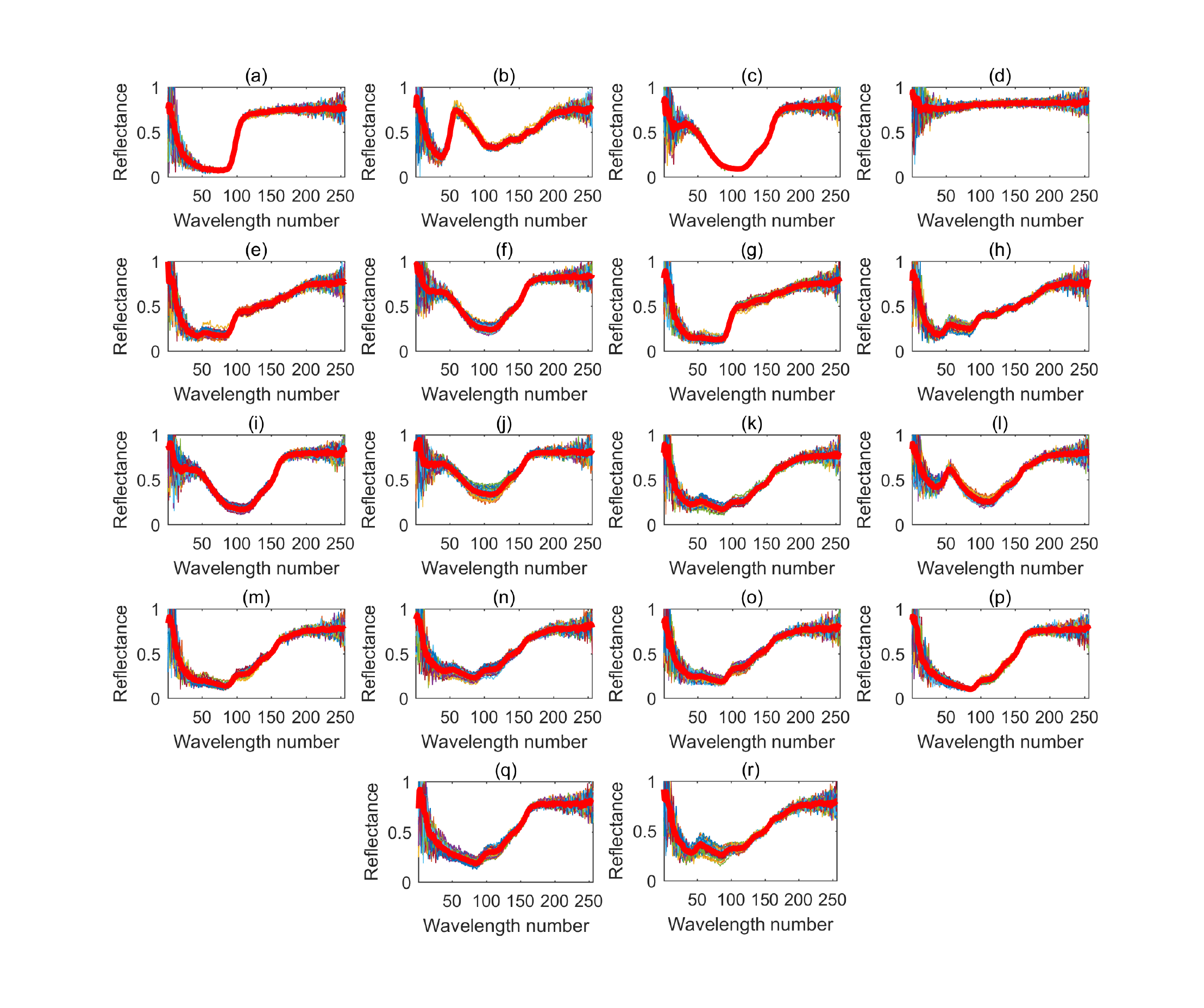}}
  \caption{Spectral curves of Scene \uppercase\expandafter{\romannumeral2}: (a) to (d) Pure colored quartz sands, red, green, blue, white in turn. (e)-(r) Fourteen intimate mixturesm, Mixture 1-Mixture 14 in turn.}
\label{fig:curves2}
\end{figure*}

\begin{figure*}
  \centering
  \centerline{\includegraphics[width=15cm]{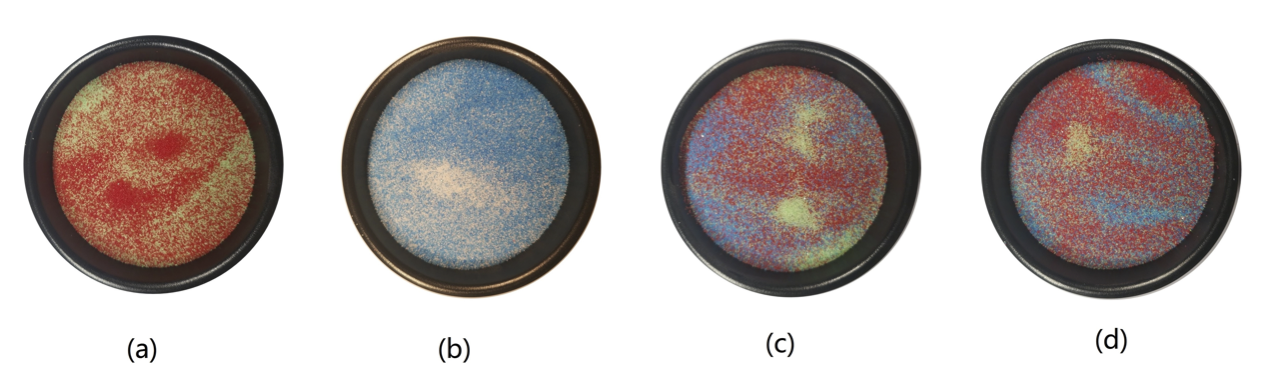}}
  \caption{Diagram of Scene \uppercase\expandafter{\romannumeral2} with quartz sands mixed with four spatial patterns (high resolution RGB images).}
\label{fig:scene2b}
\end{figure*}

In the second scene, we mimicked the intimate mixture case by using four different colors of quartz sands. The diameter of the  granules was approximately {$0.3mm$}, which was controlled by  filtering through  55-mesh and 60-mesh sieves. Using sands with a uniform size allowed us to focus on the volume fractions by ruling out the factors such as the cross-sectional areas of the granules. We first prepared fourteen mixtures where the sands were uniformly distributed as shown in Fig.~\ref{fig:scene2}.  The percentage of each color of  sands was considered as the fractional abundances with the value reported in Tab.~\ref{tab:standard2}.  The spectral curves of the endmembers and the mixtures are illustrated in Fig.~\ref{fig:curves2}. Thereafter, we prepared four mixtures with spatial patterns, as shown in~Fig.~\ref{fig:scene2b}. Unlike in the other experiments where abundances were given on the basis of the material composition,   in this experiment, the material composition of each pixel couldnot be known immediately at this setting. Therefore, we  captured high-resolution RGB images, and then analyzed the percentage of each colored sand in a low-resolution hyperspectral image with the help of the associated resolution RGB image\footnote{We captured hyperspectral images along with their associated high-resolution RGB images, and aligned them by the preset marks. Consequently, a high resolution RGB region could be linked  to a hyperspectral pixel using the spatial resolution ratio. Then the proportion of the materials in a hyperspectral pixel can be evaluated by analyzing the RGB region.}. This setting allowed to test algorithms that took into account the spatial correlation.
\begin{table} [!t]
   \footnotesize \centering
   \caption{\small Abundance ground-truth of Scene \uppercase\expandafter{\romannumeral2}.}
  \begin{tabular}{c|cccc}
  \hline  \hline
& Red & Green & Blue & White \\
  \hline
  Mixture 1 & $50\%$ & $50\%$ & $0$ & $0$\\
  Mixture 2 & $0$ & $0$ & $50\%$ & $50\%$\\
  Mixture 3 & $70\%$ & $30\%$ & $0$ & $0$\\
  Mixture 4 & $30\%$ & $70\%$ & $0$ & $0$\\
  Mixture 5 & $0$ & $0$ & $70\%$ & $30\%$\\
  Mixture 6 & $0$ & $0$ & $30\%$ & $70\%$\\
  Mixture 7 & $33.33\%$ & $33.33\%$ & $33.33\%$ & $0$\\
  Mixture 8 & $0$ & $33.33\%$ & $33.33\%$ & $33.33\%$\\
  Mixture 9 & $50\%$ & $20\%$ & $30\%$ & $0$\\
  Mixture 10 & $25\%$ & $25\%$ & $25\%$ & $25\%$\\
  Mixture 11 & $40\%$ & $20\%$ & $20\%$ & $20\%$\\
  Mixture 12 & $50\%$ & $0$ & $50\%$ & $0$\\
  Mixture 13 & $33.33\%$ & $0$ & $33.33\%$ & $33.33\%$\\
  Mixture 14 & $20\%$ & $40\%$ & $20\%$ & $20\%$\\
  \hline\hline
  \end{tabular}
\label{tab:standard2}
\vspace{-1mm}
  \end{table}

\subsubsection{Scene \uppercase\expandafter{\romannumeral3}}

\begin{figure}
  \centering
  {\includegraphics[width=9cm]{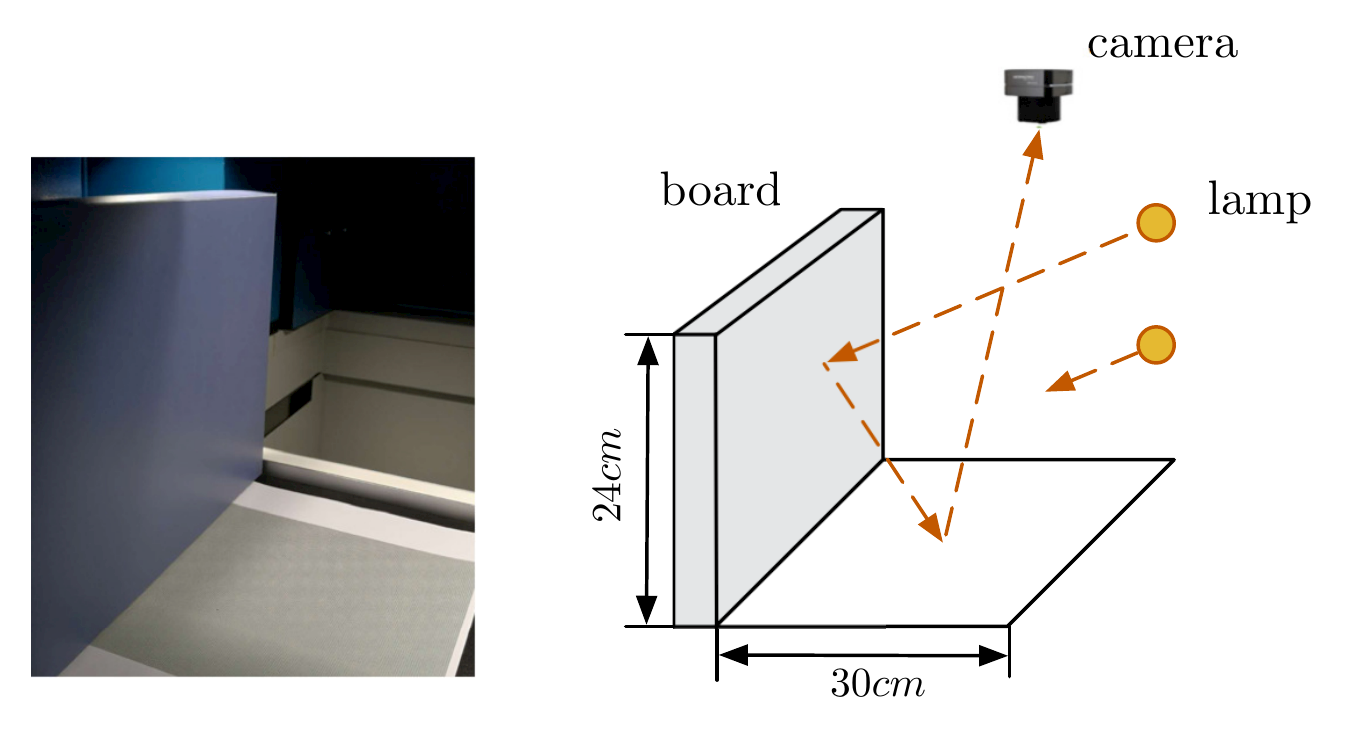}}
  \caption{Photo (left) and diagram (right) of Scene III.}
\label{fig:scene3}
\end{figure}

\begin{figure*}
  \centering
  \centerline{\includegraphics[width=20cm]{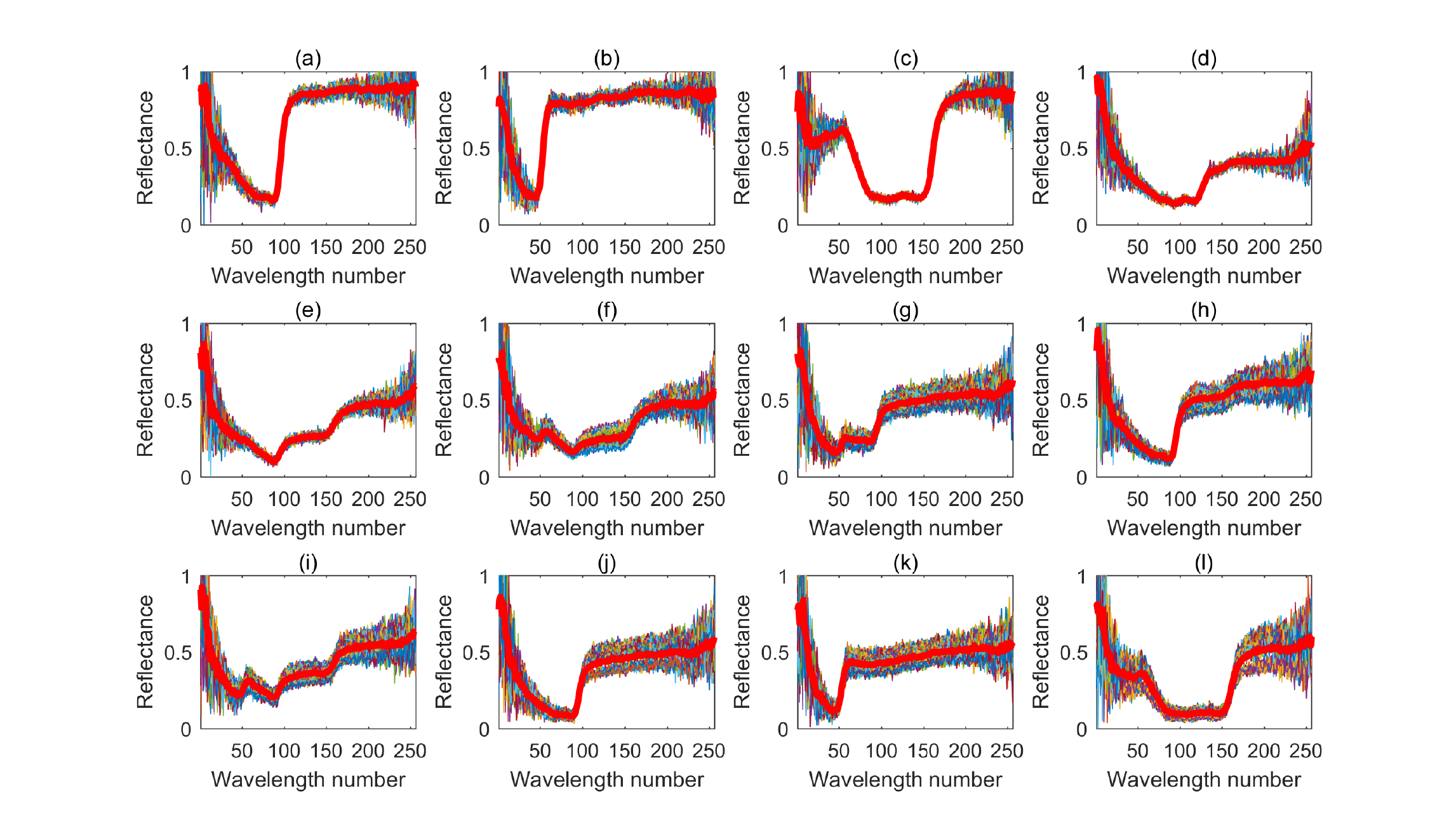}}
  \caption{Spectral curves of Scene \uppercase\expandafter{\romannumeral3}: (a) to (d) Four endmembers, magenta, yellow, cyan, board in turn.(e)-(l) Eight mixtures, Mixture 1-Mixture 8 in turn.}
\label{fig:curves3}
\end{figure*}

In this scene, we aimed to create an environment in which the second-order reflection exists. The color squares prepared in Scene I were used. Further, a plastic board was set beside the printed squares to reflect the light. A diagram of this scene is shown in Fig.~\ref{fig:scene3}. \cred{Eight mixtures were captured in this scene, including five mixtures with reflection, and three pure materials with reflection.} The mixture fractions are reported in Tab.~\ref{tab:standard3}, and the spectral curves of the endmembers and the mixtures are illustrated in Fig.~\ref{fig:curves3}.


\begin{table} [!t]
   \footnotesize \centering
   \caption{\small Abundance ground-truth of Scene \uppercase\expandafter{\romannumeral3}.}
 \vspace{1mm}
  \begin{tabular}{c|ccc}
  \hline\hline
     & Magenta & Yellow & Cyan \\
  \hline
  Mixture 1 & $50.00\%$ & $0$ & $50.00\%$\\
  Mixture 2 & $22.22\%$& $22.22\%$ & $55.56\%$\\
  Mixture 3 & $50.00\%$ & $50.00\%$ & $0$\\
  Mixture 4 & $88.89\%$ & 0 & $11.11\%$\\
  Mixture 5 & $33.33\%$ & $33.33\%$ & $33.33\%$\\
  Mixture 6 & $1$ & 0 & $0$\\
  Mixture 7 & $0$ & 1 & $0$\\
  Mixture 8 & $0$ & 0 & $1$\\
  \hline\hline
  \end{tabular}
\label{tab:standard3}
\vspace{1mm}
  \end{table}

\section{Algorithm tests with our dataset}

In this section, we apply several typical unmixing algorithms to our dataset and examine the unmixing results. Note that the purpose of this application is to check how some existing algorithms perform with these data, rather than to provide an optimal algorithm.  As it is straightforward to extract the endmember from the pure materials, we particularly focus on the unmixing with known endmembers and calculate the root mean square error (RMSE), namely,
\begin{equation}\label{eq.rmse}
  {\rm RMSE}=\sqrt{\frac{1}{NR}\sum_{i=1}^{N}\big\|  \balpha_{i}-\widehat{\balpha}_{i}\big\|}
\end{equation}
 to evaluate the abundance estimation performance of the above unmixing algorithms, where $N$ is the total number of pixels, $\balpha_{i}$ and $\widehat{\balpha}_{i}$ denote the true and estimated abundance vectors of the $i$th pixel. Endmember extraction is also tested in the linearly mixed Scene~I, as to be seen later.

\subsection{Results of Scene \uppercase\expandafter{\romannumeral1}}
\subsubsection{Abundance estimation results of Scene \uppercase\expandafter{\romannumeral1}}

\begin{figure}[!th]
  \centering
  \centerline{\includegraphics[width=10cm]{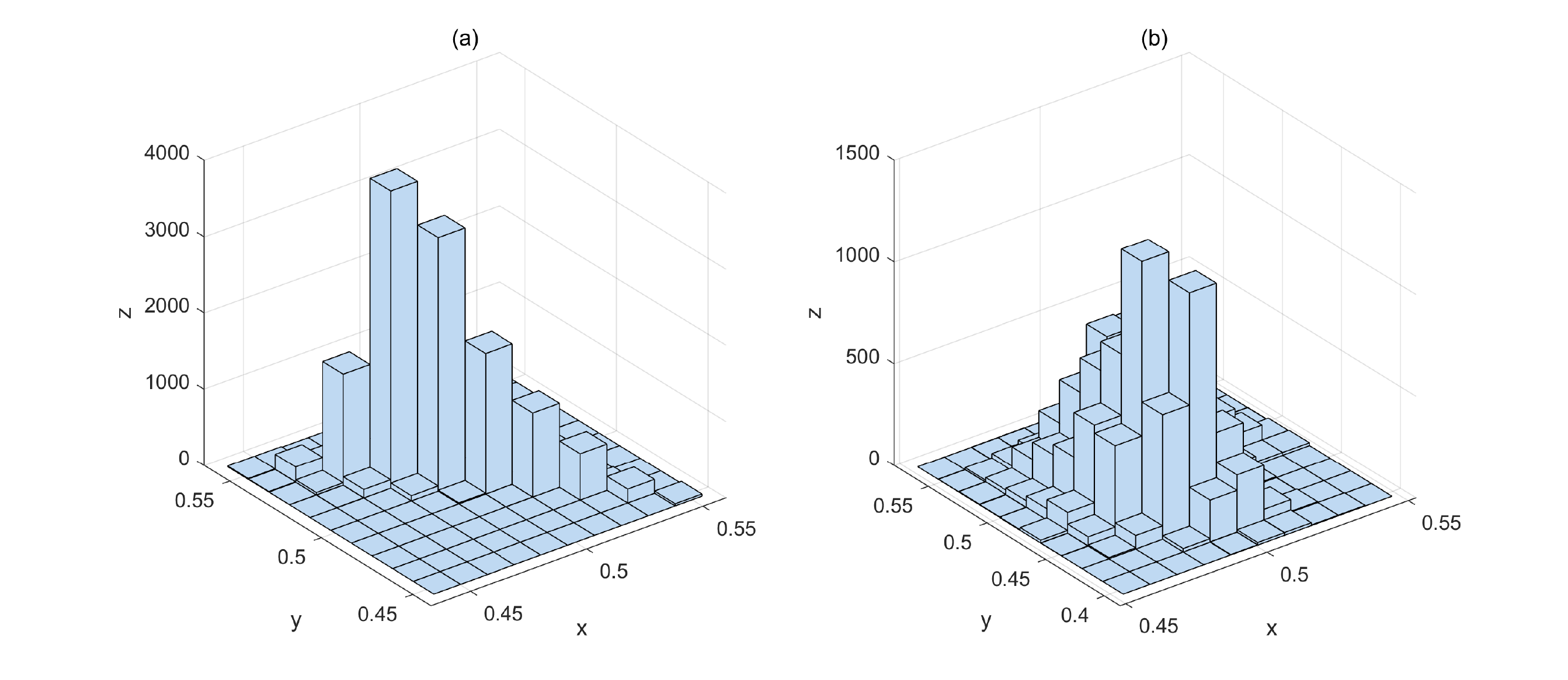}}
  \caption{The histogram of the estimates for Mixture 2 of Scene \uppercase\expandafter{\romannumeral1}: (a) FCLS, (b) NCLS,  \cred{$x$}-axis is the abundance of endmember1 (yellow), $y$-axis is the abundance of endmember2 (cyan), and $z$-axis is the number of pixels in each interval.}
\label{fig:hist1}
\end{figure}

We extracted the endmembers from the pure color blocks in Fig.~\ref{fig:scene1}, and then applied the fully constrained least square method (FCLS), and the nonnegative constrained least square method (NCLS) to the mixtures. The FCLS method relies on the linear mixture model and minimizes the least-square error, subject to the non-negativity and the sum to one constraints. By relaxing the sum-to-one constraint, the solution leads to the NCLS method.  The mean estimates and RMSE results of FCLS and NCLS are showed in Tab.~\ref{tab:result1}. We observe that both the algorithms resulted in RMSE values of the order of $10^{-2}$ and the two algorithms exhibited a comparable performance. The histograms of the estimates of Mixture 2 are shown in~Fig.~\ref{fig:hist1}.

\begin{table*} [!h]
   \footnotesize \centering
   \caption{\small Abundance estimation results of Scene \uppercase\expandafter{\romannumeral1}. Boldface number denote the lowest RMSEs.}
    \vspace{-1mm}
  \begin{tabular}{c|cccc|cccc}
  \hline
  \hline
     &\multicolumn{4}{|c|}{FCLS}&\multicolumn{4}{c}{NCLS}\\
  \hline
     & Magenta & Yellow & Cyan & RMSE & Magenta & Yellow & Cyan & RMSE \\
  \hline
  Mixture 1 & $59.51\%$ & $40.33\%$ & 0.16\% & $0.0803$ & $58.48\%$ & $40.47\%$ & 0.34\% & $\textbf{0.0749}$\\

  Mixture 2 & 0.03\%& $49.00\%$ & $50.97\%$ & $\textbf{0.0155}$ & 0& $49.47\%$ & $47.78\%$ & $0.0227$ \\

  Mixture 3 & $43.58\%$ & $2.23\%$ & $54.19\%$ & $\textbf{0.0493}$ & $42.65\%$ & $3.22\%$ & $51.60\%$ & $0.0518$\\

  Mixture 4 & $87.11\%$ & $1.66\%$ & $11.23\%$ & $0.0244$ & $87.52\%$ & $1.22\%$ & $11.72\%$ & $\textbf{0.0237}$\\

  Mixture 5 & $8.99\%$ & $0.73\%$ & $90.28\%$ & $\textbf{0.0198}$ & $8.28\%$ & $1.45\%$ & $87.03\%$ & $0.0290$\\

  Mixture 6 & $0.27$\% & $90.20\%$ & $9.53\%$ & $\textbf{0.0239}$ & $0.04$\% & $90.38\%$ & $9.48\%$ & $0.0240$\\

  Mixture 7 & $35.94\%$ & $29.85\%$ & $34.20\%$ & $0.0307$ & $35.09\%$ & $30.77\%$ & $32.15\%$ & $\textbf{0.0263}$\\

  Mixture 8 & $21.69\%$ & $20.25\%$ & $58.06\%$ & $0.0299$ & $20.19\%$ & $21.84\%$ & $54.47\%$ & $\textbf{0.0275}$\\

  Mixture 9 & $56.13\%$ & $20.76\%$ & $23.11\%$ & $0.0234$ & $55.42\%$ & $21.52\%$ & $21.39\%$ & $\textbf{0.0221}$\\

  Mixture 10 & $21.20\%$ & $56.11\%$ & $22.70\%$ & $\textbf{0.0238}$ & $21.50\%$ & $55.78\%$ & $23.43\%$ & $0.0244$\\
  \hline
  \hline
  \end{tabular}
\label{tab:result1}
\vspace{-1mm}
  \end{table*}

\subsubsection{Endmember extraction results of Scene \uppercase\expandafter{\romannumeral1}} We also applied several typical endmember extraction algorithms to the mixture shown in Figs.~\ref{fig:scene1}(b-k) and chose two widely used metrics, namely spectral angle distance (SAD) and spectral information divergence (SID), to evaluate the similarity between a real endmember and its estimate. The following methods we tested:  VCA~\cite{nascimento2005does}, N-FINDR~\cite{winter1999n}, MVC-NMF~\cite{miao2007endmember}, MVSA~\cite{Li2015}, and SPICE~\cite{zare2007sparsity}.  Among these algorithms,  the VCA and N-FINDR required the existence of pixels of pure materials, and the rest of them did not impose this requirement. Note that the scenes shown in Figs.~\ref{fig:scene1}(b-k) did not in fact contain pure color blocks.

The SAD and SID between the real endmember spectral signature $\bm$ and its estimate $\widehat{\bm}$ are respectively given by:
   ${\rm SAD}=\cos^{-1}\left(\frac{\bm^{\top}\widehat{\bm}}{\|\bm\|\|\widehat{\bm}\|}\right)$
and
$    {\rm SID}(\bm|\widehat{\bm})=\sum_{j}p_{j}\log\left(\frac{p_{j}}{\widehat{p}_{j}}\right)$
with the probability distribution vector with each endmember signature $\mathbf{p}=\frac{\bm}{\cb{1}^\top\bm}$, and its estimate $\mathbf{\widehat{p}}=\frac{\widehat{\bm}}{\cb{1}^\top\widehat{\bm}}$. The SAD and SID results of different endmember extraction methods are shown in Tab.~\ref{tab:end_result1}.
%
%

\begin{table*} [!h]
   \footnotesize \centering
   \caption{\small Endmember extraction results of Scene \uppercase\expandafter{\romannumeral1}. Boldface numbers denote the lowest SADs and SIDs.}
    \vspace{-1mm}
  \begin{tabular}{c|ccccc|ccccc}
  \hline
  \hline
     &\multicolumn{5}{|c|}{SAD}&\multicolumn{5}{c}{SID}\\
  \hline
     & VCA & N-FINDR & MVC-NMF & MVSA & SPICE & VCA & N-FINDR & MVC-NMF & MVSA & SPICE \\
  \hline
  Mixture 1 & 10.1101 & 11.2335 & 10.1722 & 10.1101 & \textbf{9.9481} & 0.0527 & 0.0676 & 0.0529 & 0.0527 & \textbf{0.0517}\\

  Mixture 2 & 15.0557 & 16.0560 & 15.1569 & 15.0557 & \textbf{15.0027} & 0.1215 & 0.1231 & 0.1231 & 0.1215 & \textbf{0.1209}\\

  Mixture 3 & 14.1272 & 15.2872 & 14.2790 & 14.1272 & \textbf{14.1095} & 0.1037 & 0.1076 & 0.1056 & 0.1037 & \textbf{0.1034}\\

  Mixture 4 & 17.5804 & 20.5323 & 17.8193 & 17.5804 & \textbf{17.5681} &0.2274 & \textbf{0.1985} & 0.2324 & 0.2274 & 0.2274\\

  Mixture 5 & 18.6483 & 19.2550 & 18.6473 & 18.6483 & \textbf{18.6134} & 0.2358 & 0.2512 & \textbf{0.2349} & 0.2358 & 0.2358\\

  Mixture 6 & 19.1086 & 21.3061 & 19.4464 & 19.1086 & \textbf{19.0999} & 0.2598 & \textbf{0.2141} & 0.2711 & 0.2598 & 0.2596\\

  Mixture 7 & 15.7100 & 16.3647 & 16.4140 & 16.3326 & \textbf{15.4834} & 0.1396 & 0.1389 & 0.1409 & 0.1405 & \textbf{0.1364}\\

  Mixture 8 & \textbf{16.4893} & 17.1885 & 17.4584 & 18.2166 & 16.6785 & \textbf{0.1301} & 0.1312 & 0.1326 & 0.1494 & 0.1329\\

  Mixture 9 & 16.0829 & 17.2970 & 17.8173 & 18.6586 & \textbf{15.9036} & 0.1591 & 0.1467 & 0.1923 & 0.1870 & \textbf{0.1533}\\

  Mixture 10 &\textbf{16.0358} & 17.5287 & 17.9511 & 18.1627 & 16.3448 & \textbf{0.1668} & 0.1624 & 0.1826 & 0.1890 & 0.1685\\
  \hline
  \hline
  \end{tabular}
\label{tab:end_result1}
\vspace{-1mm}
  \end{table*}

\subsection{Results of Scene \uppercase\expandafter{\romannumeral2}}
\subsubsection{Unmixing results of the spatially uniform mixtures}
The FCLS and NCLS were firstly applied to Scene \uppercase\expandafter{\romannumeral2}, with the mean estimates and RMSE results reported in the first and the second columns of Tab.~\ref{tab:result2}. We observe that in this scene both of these algorithms led to high estimation RMSEs, possibly caused by the fact that the intimate interaction of granules resulted in nonlinear mixture effects.  Considering this,  we then applied the nonlinear unmixing algorithm K-Hype to the data. The K-Hype algorithm models the observed pixel by a linear mixture and a nonlinear fluctuation of spectral signatures, and it is a kernel based algorithm to perform the unmixing, see~\cite{chen2013nonlinear} for more details.  We used the Gaussian kernel, $\kappa_{\rm_nlin}(\bm_{\lambda_{\ell}},\bm_{\lambda_{p}})=\exp\left(-\frac{\|\bm_{\lambda_{\ell}}-\bm_{\lambda_{p}}\|^2}{2\sigma^2}\right)$ with $\sigma =2$.
The following methods were also tested $p$-linear ($p=4$)~\cite{marinoni2015novel}, MLM~\cite{heylen2016multilinear}, Hapke model~\cite{hapke1981bidirectional}.
The obtained results are reported in the third column of  Tab.~\ref{tab:result2} and Tab.~\ref{tab:result2_2}. We observe that K-Hype and Hapke model led to reduced RMSEs for most of the mixtures. This showed the usefulness of the nonlinear unmixing.

 \begin{table*} [!th]
   \footnotesize \centering
   \caption{\small Estimation results of Scene \uppercase\expandafter{\romannumeral2}. Boldface number denote the lowest RMSEs in Tab.~\ref{tab:result2} and Tab.~\ref{tab:result2_2}.}
    \vspace{-1mm}
    {\renewcommand{\tabcolsep}{1mm}
  \begin{tabular}{c|ccccc|ccccc|ccccc}
  \hline
  \hline
     &\multicolumn{5}{|c|}{FCLS}&\multicolumn{5}{c|}{NCLS}&\multicolumn{5}{c}{K-Hype}\\
  \hline
& Red & Green & Blue & White & RMSE& Red & Green & Blue & White & RMSE& Red & Green & Blue & White & RMSE\\
  \hline
  Mixture 1 & $62.84\%$ & $37.16\%$ & $0$ & $0$ & $0.1371$& $61.59\%$ & $33.19\%$ & $0$ & $0$ & $0.1528$ & $59.11\%$ & $40.89\%$ & $0$ & $0$ & $0.0983$\\

  Mixture 2 & $0$ & $0$ & $73.00\%$ & $27.00\%$ & $0.2350$& $0$ & $0$ & $81.69\%$ & $23.90\%$ & $0.2944$ & $0$ & $0$ & $75.41\%$ & $24.59\%$ & $0.2609$\\

  Mixture 3 & $76.38\%$ & $23.62\%$ & $0$ & $0$ & $0.0763$& $75.36\%$ & $20.36\%$ & $0$ & $0$ & $0.0893$ & $69.50\%$ & $30.50\%$ & $0$ & $0$ & $\textbf{0.0334}$\\

  Mixture 4 & $49.20\%$ & $50.80\%$ & $0$ & $0$ & $0.2011$& $47.89\%$ & $46.60\%$ & $0$ & $0$ & $0.2177$ & $48.42\%$ & $51.58\%$ & $0$ & $0$ & $0.1901$\\

  Mixture 5 & $0$ & $0$ & $83.90\%$ & $16.10\%$ & $0.1453$& $0$ & $0$ & $87.69\%$ & $14.75\%$ & $0.1703$& $0$ & $0$ & $79.01\%$ & $20.99\%$ & $\textbf{0.1054}$\\

  Mixture 6 & $0$ & $0$ & $58.48\%$ & $41.52\%$ & 0.2909& $0$ & $0$ & $63.32\%$ & $39.80\%$ & $0.3242$ & $0$ & $0$ & $65.61\%$ & $34.39\%$ & $0.3626$\\

  Mixture 7 & $36.51\%$ & $16.65\%$ & $46.84\%$ & $0$ & $0.1357$ & $35.76\%$ & $13.75\%$ & $44.90\%$ & $0$ & $0.1424$ & $34.90\%$ & $28.51\%$ & $36.60\%$ & $0$ & $\textbf{0.0533}$\\

  Mixture 8 & $0$ & $37.34\%$ & $49.39\%$ & $13.27\%$ & $0.1616$ & $0$ & $43.49\%$ & $50.40\%$ & $9.37\%$ & $0.1897$ & $0$ & $36.35\%$ & $42.57\%$ & $21.09\%$ & $0.1099$\\

  Mixture 9 & $46.41\%$ & $7.10\%$ & $46.49\%$ & $0$ & $0.1313$ & $45.74\%$ & $4.70\%$ & $44.72\%$ & $0$ & $0.1336$ & $42.66\%$ & $21.45\%$ & $35.89\%$ & $0$ & $\textbf{0.0662}$\\

  Mixture 10 & $36.81\%$ & $21.49\%$ & $39.26\%$ & $2.46\%$ & $0.1555$ & $37.50\%$ & $23.94\%$ & $39.42\%$ & $0.75\%$ & $0.1625$ & $27.86\%$ & $19.51\%$ & $30.47\%$ & $22.15\%$ & $\textbf{0.0848}$\\

  Mixture 11 & $48.72\%$ & $13.94\%$ & $36.91\%$ & $0.43\%$ & $0.1471$ & $48.75\%$ & $13.94\%$ & $36.66\%$ & $0.11\%$ & $0.1476$& $37.30\%$ & $14.20\%$ & $27.61\%$ & $20.89\%$ & $0.0863$\\

  Mixture 12 & $42.25\%$ & $0$ & $57.75\%$ & $0$ & $0.0875$& $40.95\%$ & $0$ & $54.53\%$ & $0$ & $0.0832$& $45.15\%$ & $0$ & $54.85\%$ & $0$ & $\textbf{0.0586}$\\

  Mixture 13 & $43.26\%$ & $0$ & $52.81\%$ & $3.94\%$ & $0.2182$ & $43.92\%$ & $0$ & $53.78\%$ & $3.26\%$ & $0.2252$ & $29.29\%$ & $0$ & $39.20\%$ & $31.51\%$ & $\textbf{0.0867}$\\

  Mixture 14 & $32.07\%$ & $29.26\%$ & $37.41\%$ & $12.60\%$ & $0.1594$ & $32.31\%$ & $30.05\%$ & $37.30\%$ & $0.51\%$ & $0.1608$& $24.77\%$ & $26.11\%$ & $29.29\%$ & $19.82\%$ & $0.1135$\\
  \hline
  \hline
  \end{tabular}}
\label{tab:result2}
\vspace{-1mm}
  \end{table*}
\begin{table*} [!th]
   \footnotesize \centering
   \caption{\small Estimation results of Scene \uppercase\expandafter{\romannumeral2}. Boldface number denote the lowest RMSEs in Tab.~\ref{tab:result2} and Tab.~\ref{tab:result2_2}.}
    \vspace{-1mm}
    {\renewcommand{\tabcolsep}{1mm}
  \begin{tabular}{c|ccccc|ccccc|ccccc}
  \hline
  \hline
     &\multicolumn{5}{|c|}{$p$-linear}&\multicolumn{5}{c|}{MLM}&\multicolumn{5}{c}{Hapke}\\
  \hline
& Red & Green & Blue & White & RMSE& Red & Green & Blue & White & RMSE& Red & Green & Blue & White & RMSE\\
  \hline
  Mixture 1 & $62.72\%$ & $20.83\%$ & $0$ & $0$ & $0.2333$
  & $63.14\%$ & $36.86\%$ & $0$ & $0$ & $0.1404$
  & $49.25\%$ & $50.75\%$ & $0$ & $0$ & $\textbf{0.0887}$\\

  Mixture 2 & $0$ & $0$ & $73.00\%$ & $26.99\%$ & $0.2350$
  & $0$ & $0$ & $84.25\%$ & $15.75\%$ & $0.3454$
   & $0$ & $0$ & $40.05\%$ & $59.05\%$ & $\textbf{0.1181}$\\

  Mixture 3 & $76.25\%$ & $9.63\%$ & $0$ & $0$ & $0.1592$
  & $77.02\%$ & $22.98\%$ & $0$ & $0$ & $0.0824$
   & $71.15\%$ & $28.85\%$ & $0$ & $0$ & $0.0999$\\

  Mixture 4 & $49.05\%$ & $34.26\%$ & $0$ & $0$ & $0.2973$
  & $49.31\%$ & $50.69\%$ & $0$ & $0$ & $0.2025$
   & $31.11\%$ & $68.89\%$ & $0$ & $0$ & $\textbf{0.0798}$\\

  Mixture 5 & $0$ & $0$ & $83.77\%$ & $16.03\%$ & $0.1451$
  & $0$ & $0$ & $88.59\%$ & $11.41\%$ & $0.1900$
  & $0$ & $0$ & $58.33\%$ & $41.67\%$ & $0.1454$\\

  Mixture 6 & $0$ & $0$ & $58.45\%$ & $41.50\%$ & $0.2908$
  & $0$ & $0$ & $67.10\%$ & $32.90\%$ & $0.3777$
  & $0$ & $0$ & $22.98\%$ & $77.02\%$ & $\textbf{0.0905}$\\

  Mixture 7 & $35.27\%$ & $16.22\%$ & $25.72\%$ & $0$ & $0.1278$
  & $37.93\%$ & $17.29\%$ & $44.78\%$ & $0$ & $0.1292$
   & $34.77\%$ & $36.54\%$ & $28.69\%$ & $0$ & $0.0922$\\

  Mixture 8 & $0$ & $37.20\%$ & $49.36\%$ & $13.19\%$ & $0.1618$
   & $0$ & $43.75\%$ & $50.50\%$ & $5.75\%$ & $0.2055$
    & $0$ & $30.94\%$ & $28.20\%$ & $40.86\%$ & $\textbf{0.0878}$\\

  Mixture 9 & $45.20\%$ & $7.26\%$ & $26.31\%$ & $0$ & $0.1002$
   & $48.36\%$ & $6.78\%$ & $44.86\%$ & $0$ & $0.1254$
    & $50.80\%$ & $22.06\%$ & $27.14\%$ & $0$ & $0.0863$\\

  Mixture 10 & $36.64\%$ & $21.37\%$ & $37.47\%$ & $2.50\%$ & $0.1521$
   & $37.29\%$ & $23.03\%$ & $39.16\%$ & $0.52\%$ & $0.1621$
   & $23.89\%$ & $24.51\%$ & $18.53\%$ & $33.07\%$ & $0.0900$\\

  Mixture 11 & $48.32\%$ & $13.93\%$ & $30.90\%$ & $0.46\%$ & $0.1340$
   & $49.32\%$ & $14.26\%$ & $36.28\%$ & $0.14\%$ & $0.1469$
   & $39.79\%$ & $19.19\%$ & $16.80\%$ & $24.22\%$ & $\textbf{0.0851}$\\

  Mixture 12 & $41.04\%$ & $0$ & $36.10\%$ & $0$ & $0.1307$
  & $44.10\%$ & $0$ & $55.90\%$ & $0$ & $0.0729$
  & $54.81\%$ & $0$ & $45.19\%$ & $0$ & $0.0806$\\

  Mixture 13 & $42.49\%$ & $0$ & $51.18\%$ & $3.79\%$ & $0.2142$
   & $43.22\%$ & $0$ & $52.52\%$ & $4.26\%$ & $0.2162$
   & $33.45\%$ & $0$ & $27.57\%$ & $38.98\%$ & $0.0873$\\

  Mixture 14 & $31.75\%$ & $28.89\%$ & $33.69\%$ & $1.42\%$ & $0.1512$
  & $32.39\%$ & $29.97\%$ & $37.01\%$ & $0.63\%$ & $0.1595$
  & $19.97\%$ & $35.32\%$ & $17.76\%$ & $26.94\%$ & $\textbf{0.0849}$\\
  \hline
  \hline
  \end{tabular}}
\label{tab:result2_2}
\vspace{-1mm}
  \end{table*}

\subsubsection{Unmixing results of the mixtures with spatial pattern}

\begin{figure*}[!t]
  \centering
  \centerline{\includegraphics[width=15cm]{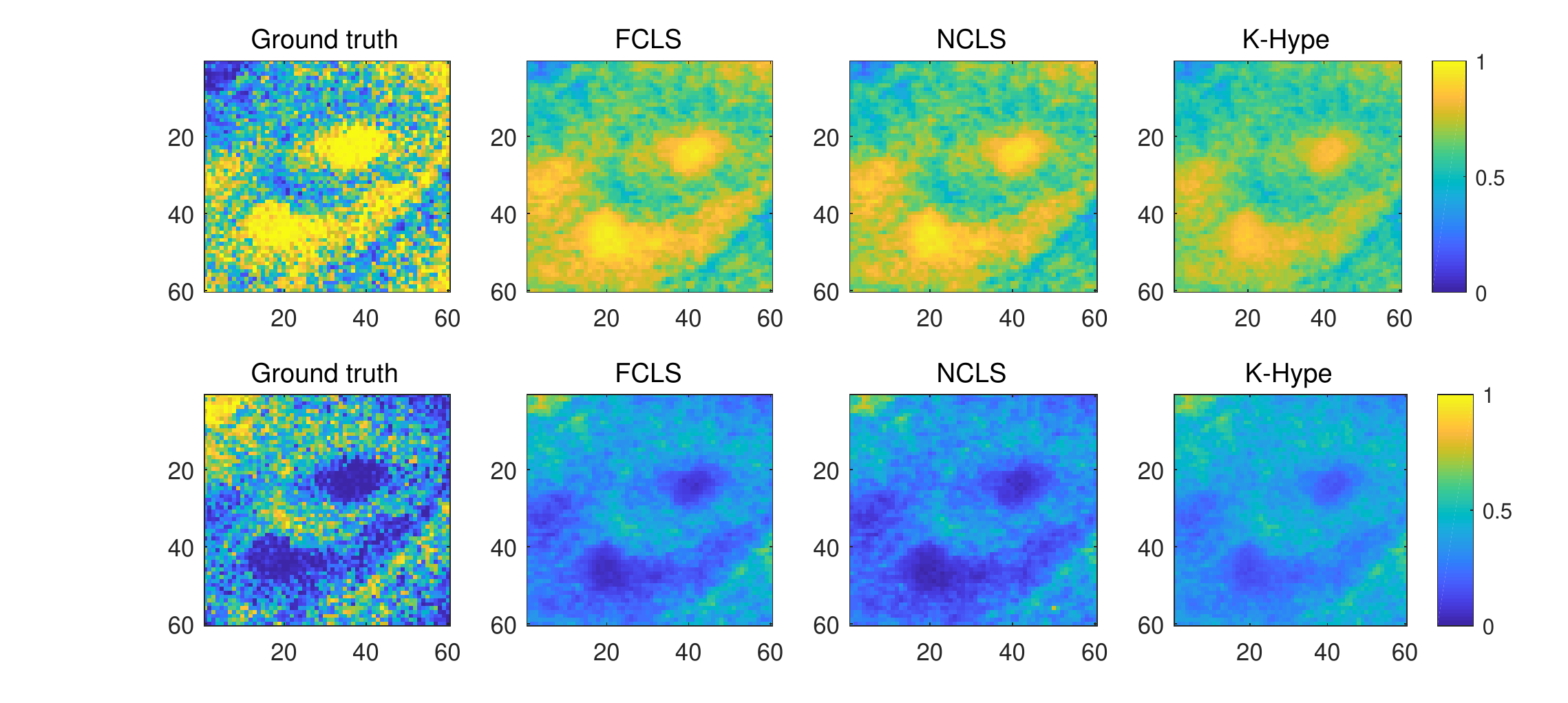}}
  \caption{Abundance maps of the mixture with spatial pattern-A in Scene II. Rows: two materials. Columns: ground-truth, results of FCLS, NCLS, K-Hype, with ${\rm RMSE} = 0.2333, 0.2387$ and 0.2242 respectively.}
\label{fig:mixA}
\end{figure*}

\begin{figure*}[!t]
  \centering
  \centerline{\includegraphics[width=15cm]{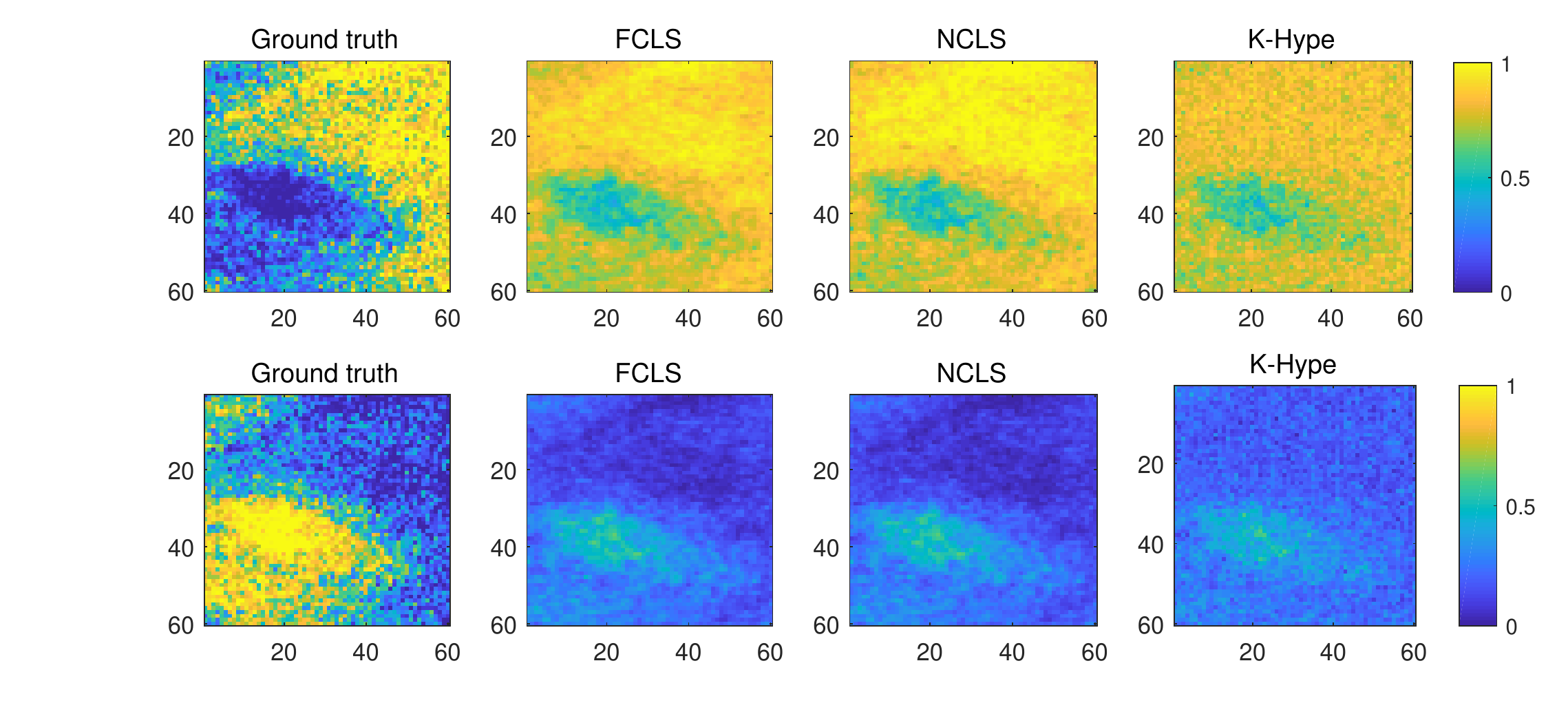}}
  \caption{Abundance maps of the mixture with spatial pattern-B in Scene II. Rows: two materials. Columns: ground-truth, results of FCLS, NCLS, K-Hype, with ${\rm RMSE} = 0.3522, 0.3682$ and 0.3543 respectively.}
\label{fig:mixB}
\end{figure*}

\begin{figure*}[!t]
  \centering
  \centerline{\includegraphics[width=15cm]{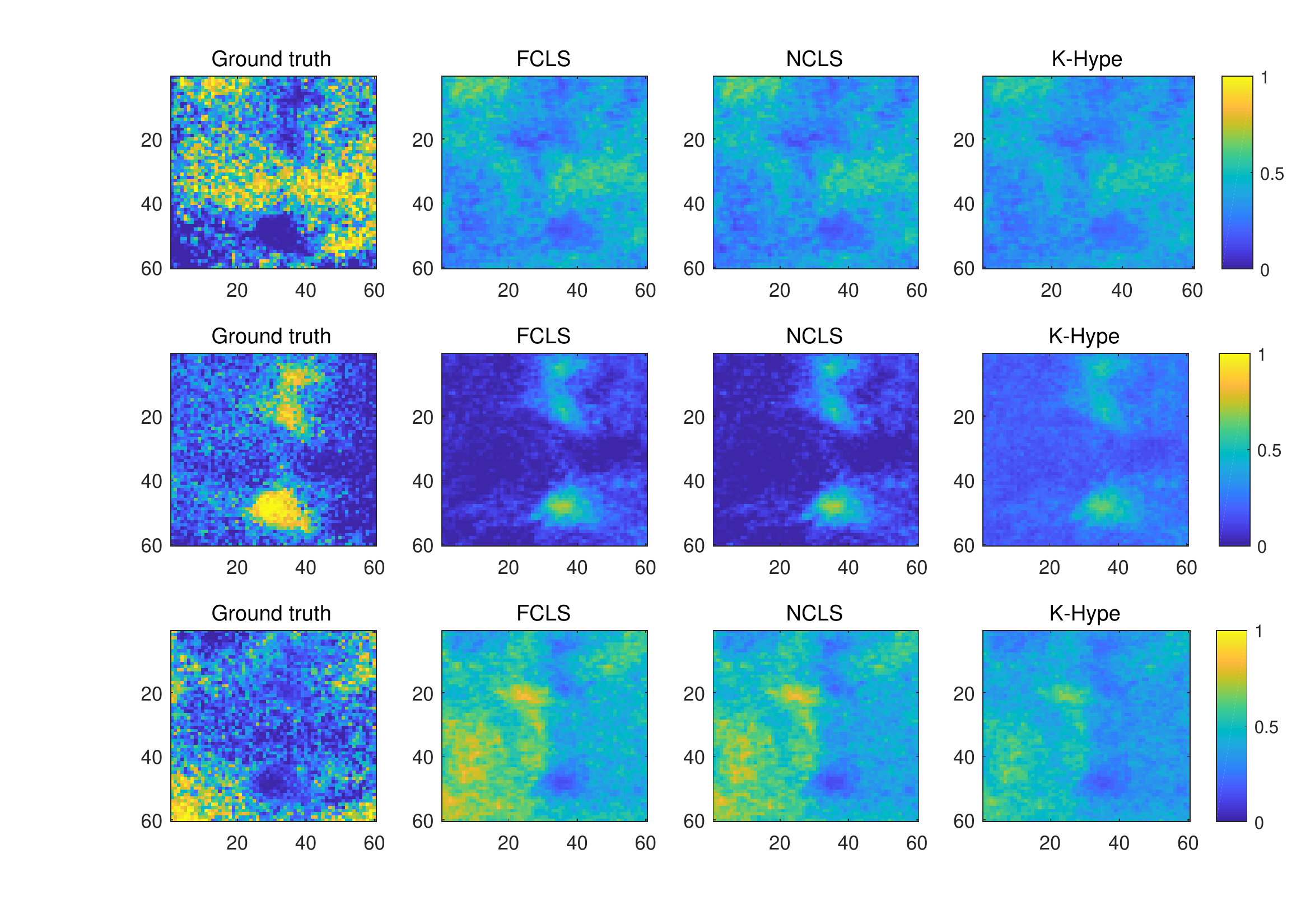}}
  \caption{Abundance maps of the mixture with spatial pattern-C in Scene II. Rows: three materials. Columns: ground-truth, results of FCLS, NCLS, K-Hype, with ${\rm RMSE} = 0.2689,  0.2663$ and 0.2396 respectively.}
  \label{fig:mixC}
\end{figure*}

\begin{figure*}[!t]
  \centering
  \centerline{\includegraphics[width=15cm]{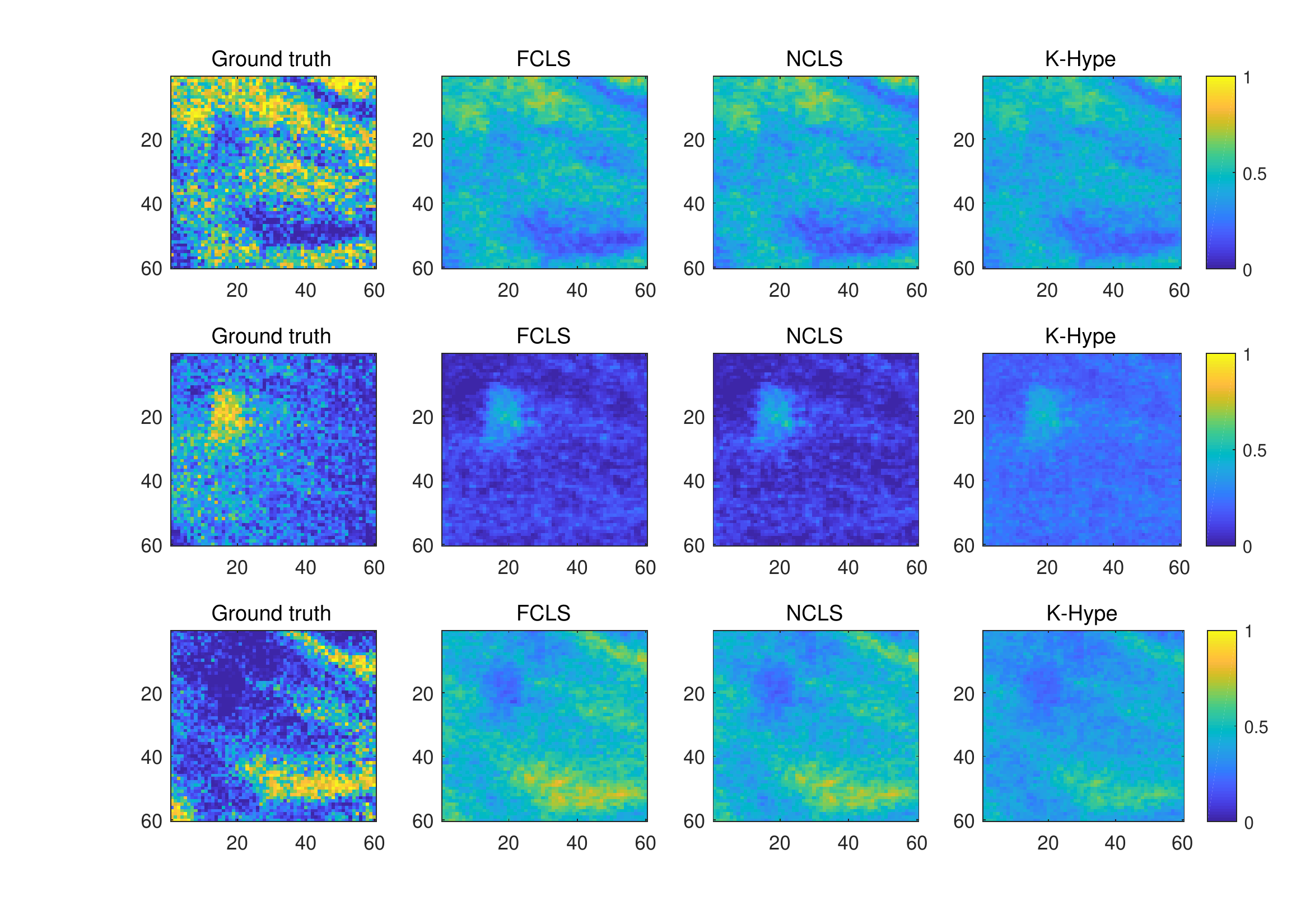}}
  \vspace{-2mm}
  \caption{Abundance maps of the mixture with spatial pattern-D in Scene II. Rows: three materials. Columns: ground-truth, results of FCLS, NCLS, K-Hype, with ${\rm RMSE} = 0.2573, 0.2553$ and 0.2235 respectively.}
  \label{fig:mixD}
\end{figure*}

The true abundance maps of these four mixtures (clipped center square regions) are shown in the first columns in Figs.~\ref{fig:mixA} to~\ref{fig:mixD}.  The abundance maps estimated using the FCLS, NCLS, and K-Hype algorithms are shown alongside. We observe that the general spatial patterns of the estimated maps were consistent with those of the ground-truth reference; however, the RMSEs are notable. The K-Hype algorithm showed better RMSEs than the other two algorithms in mixtures of patterns A, C, and D. A further study on these mixtures is necessary.


\subsection{Results of Scene \uppercase\expandafter{\romannumeral3}}
In this scene, the spectra of the pure printed colors and that of the board are used as the endmembers. Then, the unmixing algorithms were executed to estimate the abundances of the five mixtures.  The mean estimates are shown in Tab.~\ref{tab:result3}. We observe that the board significantly  contributed to the spectra because of to its strong reflection. As it was difficult to estimate the contribution from the board, we normalized the sum of the abundances of the printed color to 1 and then compared it with the {ground-truth} obtained from the layouts of the checkerboards. These RMSEs are shown in Tab.~\ref{tab:result3b}. We found that K-Hype exhibited  good performance for the mixture with multiple materials.
\begin{table*} [!h]
   \footnotesize \centering
   \caption{\small Abundance estimation result of Scene \uppercase\expandafter{\romannumeral3}.}
    \vspace{-1mm}
 {\renewcommand{\tabcolsep}{0.6mm}
  \begin{tabular}{c|cccc|cccc|cccc|cccc}
  \hline
  \hline
     &\multicolumn{4}{|c|}{FCLS}&\multicolumn{4}{c|}{NCLS}&\multicolumn{4}{c|}{K-Hype}&\multicolumn{4}{c}{Bilinear}\\
  \hline
     & Magenta & Yellow & Cyan & Board& Magenta & Yellow & Cyan & Board& Magenta & Yellow & Cyan & Board& Magenta & Yellow & Cyan & Board\\
  \hline
  Mixture 1 & $8.14\%$ & $0.06\%$ & $12.21\%$& $76.59\%$
  &$14.54\%$ & $0.06\%$ & $19.65\%$& $52.19\%$
  & $19.98\%$ & $7.22\%$ & $23.44\%$& $49.36\%$
  & $13.24\%$ & $5.83\%$ & $16.69\%$& $45.55\%$
  \\
  Mixture 2 & $0.54\%$ & $4.08\%$ & $9.15\%$ & $86.23\%$
  & $4.03\%$ & $7.01\%$ & $21.22\%$ & $48.23\%$
  & $9.62\%$ & $16.75\%$ & $25.20\%$ & $48.43\%$
  & $2.86\%$ & $7.73\%$ & $18.07\%$ & $42.99\%$
  \\
  Mixture 3 & $14.35\%$ & $12.13\%$ & $0$ & $73.52\%$
  & $27.50\%$ & $14.93\%$ & $0.02\%$ & $34.41\%$
  & $26.83\%$ & $25.10\%$ & $3.11\%$ & $44.97\%$
  & $22.19\%$ & $15.14\%$ & $0.02\%$ & $30.31\%$
  \\
  Mixture 4 & $27.77\%$ & $0$ & $0.31\%$ & $71.92\%$
  & $38.89\%$ & $0$ & $1.24\%$ & $43.26\%$
  & $39.30\%$ & $7.84\%$ & $6.89\%$ & $45.96\%$
  & $36.84\%$ & $0.06\%$ & $1.27\%$ & $36.24\%$
  \\
  Mixture 5 & $3.74\%$ & $8.15\%$ & $3.49\%$ & $84.62\%$
  & $11.09\%$ & $9.85\%$ & $10.25\%$ & $51.25\%$
  & $13.50\%$ & $20.13\%$ & $15.16\%$ & $51.21\%$
  & $8.64\%$ & $10.31\%$ & $8.83\%$ & $44.95\%$
  \\
  Mixture 6 & $21.49\%$ & $0$ & $0$& $78.51\%$
  & $38.09\%$ & $0$ & $0$& $37.92\%$
  & $39.56\%$ & $8.38\%$ & $2.41\%$& $49.65\%$
  & $36.20\%$ & $0$ & $0$& $32.69\%$
  \\
  Mixture 7 & $0$ & $35.86\%$ & $0$ & $64.14\%$
  & $0.07\%$ & $47.18\%$ & $0.06\%$ & $30.94\%$
  & $3.40\%$ & $50.97\%$ & $3.52\%$ & $42.11\%$
  & $0.03\%$ & $45.78\%$ & $0.05\%$ & $25.27\%$
  \\
  Mixture 8 & $0$ & $0$ & $17.44\%$ & $82.56\%$
  & $0$ & $0$ & $43.36\%$ & $27.68\%$
  & $1.86\%$ & $8.06\%$ & $44.07\%$ & $46.01\%$
  & $0$ & $0$ & $39.56\%$ & $26.99\%$
  \\
  \hline
  \hline
  \end{tabular}}
\label{tab:result3}
\vspace{-1mm}
  \end{table*}

  \begin{table} [!th]
   \footnotesize \centering
   \caption{\small RMSE  results (without considering the board) of Scene \uppercase\expandafter{\romannumeral3}. Boldface number denote the lowest RMSEs.}
    \vspace{-1mm}
{
  \begin{tabular}{c|c|c|c|c}
  \hline
  \hline
     &{FCLS}&{NCLS}&{K-Hype}&Bilinear\\
  \hline
Mixture 1 & $0.1720$& $0.0664$&  $ 0.1107$& $\mathbf{0.0552}$\\
Mixture 2 &  $0.2897$& $0.1434$& $ \mathbf{0.1020}$& $0.1405$\\
Mixture 3 &  $0.0568$& $0.1295$&$\mathbf{0.0565}$& $0.0892$\\
Mixture 4 &  $0.0873$& $0.0778$& $0.1384$& $\mathbf{0.0735}$\\
Mixture 5 &  $0.3560$& $0.0709$ & $0.0875$& $\mathbf{0.0567}$\\
Mixture 6 &  $\mathbf{0}$& $\mathbf{0}$ & $0.1641$& $\mathbf{0}$\\
Mixture 7 &  $\mathbf{0}$& $0.0084$ & $0.0997$& $0.0054$\\
Mixture 8 &  $\mathbf{0}$& $\mathbf{0}$ & $0.1444$& $\mathbf{0}$\\
  \hline
  \hline
  \end{tabular}}
\label{tab:result3b}
\vspace{-1mm}
  \end{table}

\section{Conclusion}
In our work, we created a dataset for evaluating the performance of spectral unmixing algorithms. The dataset provides \cred{3} types of mixture scenes and \cred{36} mixtures that simulate different mixture models. Each pixel consists of 256 spectral bands with a spectral resolution of up to $0.58nm$. Both of the endmembers of pure materials and their compositions are priorly known, allowing us to testing unmixing algorithms in a quantitative and an objective manner. Typical unmixing algorithms,  including the FCLS, NCLS, K-Hype,  Bilinear, $p$-linear, MLM, and Hapke methods were applied to this dataset and led to interpretable results.  Unmixing of some complex scenes however requires some further study. In the future, we will on one hand make efforts to enrich the dataset with more settings, and on the other hand, to devise and test unmixing algorithms with these labeled data.
\bibliographystyle{IEEEbib}
\bibliography{ref}
\balance




\end{document}